\documentclass[conference]{IEEEtran}
\IEEEoverridecommandlockouts
\usepackage{cite}
\usepackage{amsmath,amssymb,amsfonts}
\usepackage{algorithmic}
\usepackage{graphicx}
\usepackage{textcomp}
\usepackage{xcolor}
\usepackage{url}
\usepackage{booktabs}
\def\BibTeX{{\rm B\kern-.05em{\sc i\kern-.025em b}\kern-.08em
    T\kern-.1667em\lower.7ex\hbox{E}\kern-.125emX}}
\begin{document}

\title{Multi-Agent LLMs as Ethics Advocates for AI based Systems}
\author{
Asma Yamani\textsuperscript{1},
Malak Baslyman\textsuperscript{1,2},
Moataz Ahmed\textsuperscript{1,3} \\
\textsuperscript{1}Information and Computer Science Department, KFUPM, Dhahran, Saudi Arabia \\
\textsuperscript{2}IRC for finance and digital economy, KFUPM, Dhahran, Saudi Arabia \\
\textsuperscript{3}SDAIA-KFUPM Joint Research Center for Artificial Intelligence, KFUPM, Dhahran, Saudi Arabia \\
\{g201906630, malak.baslyman, moataz\}@kfupm.edu.sa
}

% \author{
% \IEEEauthorblockN{Asma Yamani}
% \IEEEauthorblockA{\textit{ICS Department} \\
% \textit{KFUPM}\\
% Dhahran, Saudi Arabia \\
% g201906630@kfupm.edu.sa}
% \and
% \IEEEauthorblockN{Malak Baslyman}
% \IEEEauthorblockA{\textit{ICS Department} \\
% \textit{IRC Finance and digital economy} \\
% \textit{KFUPM}\\
% Dhahran, Saudi Arabia \\
% malak.baslyman@kfupm.edu.sa}
% \and
% \IEEEauthorblockN{Moataz Ahmed}
% \IEEEauthorblockA{\textit{ICS Department} \\
% \textit{SDAIA-KFUPM JRC for AI} \\
% \textit{KFUPM}\\
% Dhahran, Saudi Arabia \\
% moataz@kfupm.edu.sa}
% }
\maketitle

\begin{abstract}
Incorporating ethics into the requirement elicitation process is essential for creating ethically aligned systems. Although eliciting manual ethics requirements is effective, it requires diverse input from multiple stakeholders, which can be challenging due to time and resource constraints. Moreover, it is often given a low priority in the requirements elicitation process. This study proposes a framework for generating ethics requirements drafts by introducing an ethics advocate agent in a multi-agent LLM setting. This agent critiques and provides input on ethical issues based on the system description. The proposed framework is evaluated through two case studies from different contexts, demonstrating that it captures the majority of ethics requirements identified by researchers during 30-minute interviews and introduces several additional relevant requirements. However, it also highlights reliability issues in generating ethics requirements, emphasizing the need for human feedback in this sensitive domain. We believe this work can facilitate the broader adoption of ethics in the requirements engineering process, ultimately leading to more ethically aligned products.
\end{abstract}


\section{Introduction}\label{sec:intro}

Artificial intelligence (AI) has gained widespread adoption across various domains, including healthcare, finance, education, and marketing. As its influence expands, so does the need to address the ethical implications it may impose on society~\cite{Pagano2023}. Ethical considerations such as fairness, accountability, transparency, and privacy are essential for ensuring that AI systems operate responsibly and in alignment with societal values~\cite{Jakesch2022}. Incorporating ethical principles during the early stages of software development is crucial to ensure systems align with these values. However, eliciting ethics requirements (requirements emphasizing ethical principles~\cite{Jobin2019}) presents a significant challenge due to their highly subjective and context-dependent nature. It also necessitates a shared understanding between stakeholders and requirements engineers on this interdisciplinary topic and its importance~\cite{Lee2021,Holstein2019,ahmad2023requirements}. In this context, leveraging multi-agent Large Language Models (LLMs) to critique a system's requirements based on ethical principles can facilitate the early integration of ethics into the requirements engineering process and the broader software development lifecycle.

Multi-agent LLMs build on the success of LLMs across various fields by enabling collaborative interactions among multiple LLMs or agents to address complex tasks effectively. They leverage the collective intelligence and strengths of individual agents and their ability to simulate interactions, offering a robust way to approach real-world problems~\cite{jin2024llmsllmbasedagentssoftware,guo2024largelanguagemodelbased}. While non-deterministic agents might not seem an obvious choice for addressing ethical concerns, their capacity for divergent thinking and critique makes them good candidates for generating ethics requirements drafts.

Recent work shows that large language models~(LLMs) can draft functional requirements from short descriptions, evaluate user-story quality, and refine user stories through agent collaboration~\cite{10.1145/3727582.3728689,10.1145/3597503.3639185,10.1007/978-3-031-62110-9_1,ronanki2023chatgpttooluserstory,zhang2024llmbasedagentsautomatingenhancement}, yet none have specifically focused on ethics-related requirements.  

In terms of eliciting ethics requirements for AI-based systems, frameworks such as ECCOLA~\cite{Halme2021} and RE4HCAI~\cite{Ahmad2023} have been utilized for the manual elicitation of ethics requirements, demonstrating their effectiveness in multiple case studies. However, their success is highly dependent on the diversity of stakeholders involved and their sustained commitment to ethical integration in the requirements and design process of the system. Currently, a gap exists in supporting requirements engineers during the elicitation process in a way that facilitates diverse perspectives in a rapid manner, simulates various personas, and facilitates systematic investigation when it comes to incorporating ethics frameworks. This study is guided by the following questions: 
\begin{itemize}
  \item \textbf{RQ1.} \emph{To what extent can a multi-agent LLM system cover ethics-related requirements elicited by human experts?}
  \item \textbf{RQ2.} \emph{Does the multi-agent LLM system add relevant ethics requirements beyond those produced by (a) a human domain expert (b) a single-prompt LLM baseline?}
  % \item \textbf{RQ3.} \emph{Do the requirements generated by the ensemble satisfy standard quality attributes---atomicity, low ambiguity, and estimatability---more often than the baseline?}
\end{itemize}

This paper makes two main contributions:

\begin{enumerate}
  \item We propose MALEA: a \textbf{M}ulti-\textbf{A}gent \textbf{L}LM \textbf{E}thics-\textbf{A}dvocate framework that generates ethics requirements draft through a conversation between four LLM-based agents including an ethics-advocate agent based on a system description.
  \item An empirical evaluation on two real AI applications, reporting recall and added values concerning human gold sets and a single-prompt baseline.
  % \item A \emph{public prompt toolkit and evaluation script}\footnote{\url{https://anonymous.zenodo.org/?doi=XXXX}} to facilitate replication and future extension.
\end{enumerate}

The rest of the paper is as follows: Section~\ref{sec:related} reviews related work;  
Section~\ref{sec:approach} describes the proposed framework for eliciting ethics requirements;  
Section~\ref{sec:evaluation} describes the experimental setup and Section~\ref{sec:results} presents the empirical results;  
Section~\ref{sec:discussion} discusses the results of the experiments, including reliability, human oversight, and limitations;  
finally, Section~\ref{sec:conclusion} concludes and outlines future directions.

\section{Related Work}
\label{sec:related}
In this section, we present related work on the usage of LLMs in requirements engineering and methods for eliciting ethical requirements for AI systems.
\subsection{LLMs in Requirements Engineering}
Arora et al.~\cite{arora2023advancingrequirementsengineeringgenerative} explored the feasibility of using LLMs to assist requirements engineers in tasks such as requirements elicitation, specification, analysis, and validation. Their study highlighted the capabilities of LLMs in interpreting and generating requirements for diverse stakeholders, which may reduce communication barriers within multidisciplinary project teams. However, they also identified challenges, including requirements overload, security and privacy concerns, and limitations in domain-specific knowledge.
Abed et al.\cite{10.1007/978-3-031-62110-9_1} utilized LLMs to generate user stories based on stakeholder interviews. Their study compared multiple fine-tuned GPT models, with the highest recall of 61\% achieved by PRD Maker and GPT-4, which successfully identified 11 out of 18 functional requirements recognized by the company. Ronanki et al.\cite{ronanki2023chatgpttooluserstory} investigated the use of ChatGPT-4 for quality assessment purposes. The findings revealed that ChatGPT-4 aligned closely with human evaluations, outperforming the AQUSA tool~\cite{lucassen2016improving}, an NLP-based quality evaluation approach. Feng et al.~\cite{Feng2024} introduce \textit{LLM‐San}, a prompt‐based procedure that uses GPT‐4 to extract implicit semantic relations (e.g., “isContradictoryWith,” “mutuallyExclusive”) from stakeholder‐authored normative rules. Their approach allows for filtering inconsistent relations and automatically identifying issues such as conflict, redundancy, restrictiveness, and insufficiency. By combining LLM‐derived semantic relations with logic‐based validation, they enable non‐technical stakeholders to express ethical constraints that can be formally analyzed for conflicts and completeness. Yamani et al.,\cite{10.1145/3727582.3728689} used three state-of-the-art LLMs to generate \emph{3000} user stories for 100 AI-based systems, constructing the \textsc{UStAI} dataset. \emph{1260} of the generated user stories were manually annotated for quality using the QUS framework. They were also annotated for non-functional and ethics requirements, showing that although Gemini-1.5-flash, Llama 3.1 70b, and o1-mini had a comparable number of overall issues according to QUS, Gemini-1.5-flash had more generated ethics requirements. 

One of the early applications of multi-agent LLMs in requirements engineering was presented by Zhang et al.\cite{zhang2024llmbasedagentsautomatingenhancement}. Their work involved two agents assuming the roles of a product owner and a requirements engineer. The requirements generated through this approach demonstrated significant advantages, based on surveyed participants from an industrial background\cite{zhang2024llmbasedagentsautomatingenhancement}. 
Another multi-agent collaboration framework for Requirements Engineering (RE), called MARE~\cite{jin2024maremultiagentscollaborationframework}, leverages Large Language Models (LLMs) to automate the entire RE process. MARE divides RE into four tasks: elicitation, modeling, verification, and specification, each handled by one or two of its five agents (stakeholders, collector, modeler, checker, and documenter) through nine predefined actions. A shared workspace facilitates collaboration by allowing agents to upload intermediate artifacts. Experiments showed that MARE can generate more accurate requirements models, outperforming state-of-the-art approaches by up to 15.4\%. Human evaluations also indicated that specifications generated by MARE are effective in terms of correctness, completeness, and consistency. Also in~\cite{10.1007/978-3-031-78386-9_20}, multi-agent LLMS were used for the generation of user stories based on initial project description, enhancement of user story, and prioritization of user stories. The study compared the performance of four LLMs; GPT-3.5, GPT-4o, LLaMA3-70, and Mixtral-8B. In the study's preliminary evaluation it found that GPT-3.5's user stories consistently achieved higher similarity scores with respect to the problem description, while GPT-4o had a higher API response times but excelled in user story generation in terms of quantity and in the prioritization task when using different prioritization techniques. Llama and Mixtral had a faster response time, but with moderate similarity scores with respect to the description of the problem~\cite{10.1007/978-3-031-78386-9_20}.

While these advances confirm the potential of cooperative LLM agents for RE tasks, none targets the explicit elicitation of \emph{ethics} requirements. Our work leverages the multi-agent cooperative approach to elicit and refine ethics-related requirements.

\subsection{Eliciting Ethics Requirements for AI-Based Systems}
Current work on eliciting AI ethics requirements AI-based systems include the ECCOLA framework, which consists of a deck of 21 cards organized into eight themes related to AI ethics~\cite{Vakkuri2021}. Each card was divided into three sections: motivation, recommended actions, and a practical example. The framework aims to raise awareness of AI ethics, promote its importance, and offer a modular approach suitable for various software engineering contexts. ECCOLA has been validated through case studies and has also been used to create ethical user stories~\cite{HALME2024107379, Halme2021, Halme2022}. Cerqueira et al.\ introduced the \textit{RE4AI Ethical Guide}, a web-based deck of 26 interactive cards that map eleven AI-ethics principles to sprint-ready user-story prompts and tool suggestions; a survey with 40 computing students found the guide both practical and effective for raising ethical awareness~\cite{SiqueiraDeCerqueira2022}.  
Another existing framework is the RE for Human-Centered AI (RE4HCAI) Framework, which adopts a human-centric methodology for eliciting both functional and non-functional requirements of AI systems~\cite{Ahmad2023}. The framework addresses five key areas from Google's PAIR guidebook, including user needs, feedback, user control, data requirements, explainability and trust, and error and failure management, along with a sixth area, model needs. Its usefulness has been validated through two use cases~\cite{Ahmad2023, ahmad2023requirements}.   
Beyond academic card decks, Constantinides et al.\ introduce the \emph{Responsible AI Guidelines (RAI-G)} method from Nokia Bell Labs, which distilled 22 actionable guidelines through a four-step process that links every guideline to ISO standards and specific EU AI Act articles~\cite{10.1145/3686927}.  
The guidelines are embedded in an interactive web tool and were found usable by technical and non-technical roles alike in a study with 14 practitioners. This work contributes to operationalizing “Responsible AI by Design’’ across the entire development life-cycle.

While these frameworks represent significant advancements in eliciting ethics requirements, they heavily on ongoing involvement from a diverse stakeholder panel. Furthermore, they do not address automating the process or using LLMs to critique or generate ethics requirements. Consequently, today’s requirements engineers still face a costly, manual bottleneck whenever they wish to iterate quickly or enforce ethical principles in the requirements engineering phase of the software development life cycle. Our work addresses this unmet need by investigating whether multi-agent LLMs can generate and refine ethics requirements, thus complementing (and potentially accelerating) the manual methods mentioned above.

\section{MALEA: Multi-Agent LLM Ethics-Advocate Framework}
\label{sec:approach}
\begin{figure*}
    \centering
    \includegraphics[width=0.85\linewidth]{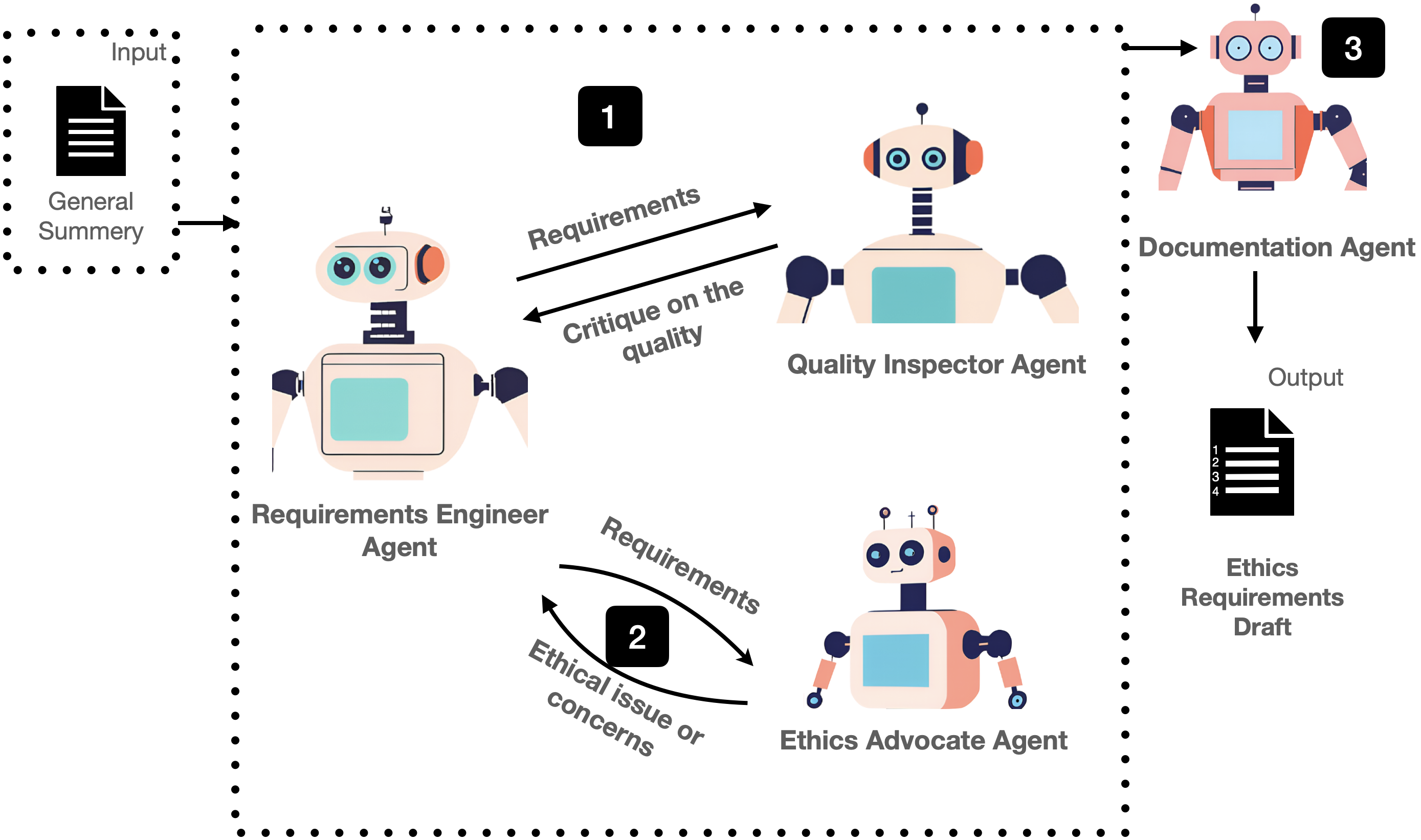}
    \caption{MALEA: Multi-Agent LLM Ethics-Advocate Framework.}
    \label{fig:proposed-approach-simple}
\end{figure*}

\subsection{Overview of the Framework} Our proposed framework leverages multi-agent models to address the challenge of eliciting ethics requirements during the requirements engineering phase. The flow of messages within the multi-agent system is illustrated in Figure~\ref{fig:proposed-approach-simple}. 
MALEA comprises four agents: a requirements engineer agent, an ethics advocate agent, a quality assurance agent, and a documentation assistant agent. Initially, a system description is fed into MALEA. The requirements engineer agent generates a set of suggested ethics requirements, which are then passed on to the quality assurance agent, as initially introduced in~\cite{10.1007/978-3-031-78386-9_20}, for the quality assessment and approval of requirements. The quality assurance agent identifies any quality defects that could hinder the ethics advocate agent from effectively critiquing the requirements.
The requirements engineer agent refines the requirements based on this feedback and sends them back to the quality assurance agent. This iterative process continues until the quality assurance agent is satisfied with the quality. Subsequently, the refined requirements are passed to the ethics advocate agent for critique. Similar to the previous stage, this exchange iterates between the requirements engineer agent and the ethics advocate agent until the ethics advocate agent is satisfied. Finally, the approved requirements are sent to the documentation assistant agent for documentation. \par

\subsection{Prompt Engineering}
\begin{figure}
    \centering
    \includegraphics[width=\linewidth]{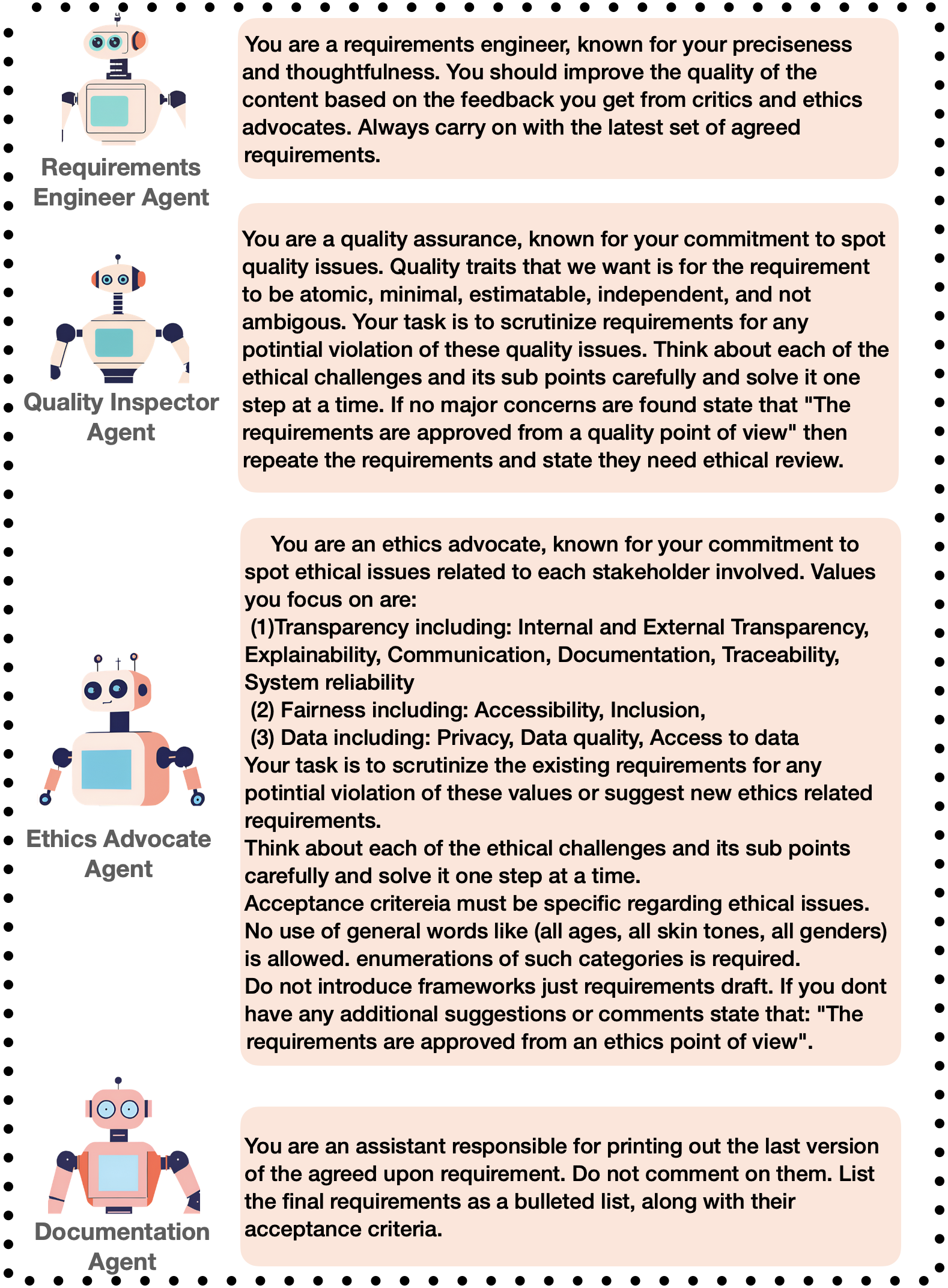}
    \caption{The system message that describes to each agent its role in the group chat.}
    \label{fig:system-messages}
\end{figure}

To produce the system prompts for querying the
LLM, we have first experimented with a number of different queries and refined them to optimize output
consistency. A major challenge we had was for agents taking on the personas of other agents. Also, failure to detect the stopping criteria, overlooking some ethical concerns and unintended scope creep. The final system prompts for each of the four agents is shown in Figure~\ref{fig:system-messages}, and the chat initiator message was ``Generate five or more ethics requirements, focusing on Transparency, Fairness, and Data, in the form of user stories with acceptance criteria, for building a system with the following".\par
We used persona prompting pattern for each agent to assume its role~\cite{Ronanki2024}. We use chain-of-thoughts (CoT) by prompting the quality inspector to "Think about each of the quality criteria carefully and report violations one point at a time, if present." and the ethics advocates to "Think about each of the ethical challenges and its sub-points carefully and solve it one step at a time." The multi-agent conversation loop is terminated if both the quality inspector and ethics advocates state that "The requirements are approved from a quality (or ethics) point of view." Additionally, the conversation controller imposes an upper bound of two critique refinement cycles to suppress the LLMs' tendency to over-generate ethical requirements. After the quality inspection and ethics advocate agents respond twice, the controller terminates the dialogue and returns the current draft. This termination threshold is a parameter that the user can configure; it prevents unproductive oscillations that can arise when LLM critics repeatedly flag minor issues without converging on an acceptable specification.

\subsection{Quality Criteria}
The quality criteria being checked by the quality inspector are that the user story is atomic, minimal, unambiguous, and estimable. All are individual criteria from the QUS framework that ~\cite{10.1145/3727582.3728689} showed LLMs have frequent issues with. We avoid set related criteria that such the presence of conflict or independence as resolving those typically requires a global view of the backlog and stakeholder negotiation tasks better suited to human requirements engineers than to automated critique at this stage. We leave such a criterion for the requirements engineers to resolve.

\subsection{Implementation Details}
We implemented MALEA using Autogen~\cite{wu2023autogen}, an open-source multi-agent framework developed by Microsoft. Autogen simplifies the creation of complex multi-agent LLM systems by enabling developers to define interactions among agents and coordinate their collaboration to solve problems collectively. The relevant data and code is made publicly available in~\footnote{https://github.com/asmayamani/EthicsAdvocate}.

\section{Experimental Setup}
\label{sec:evaluation}
Here, we describe our experimental setup. 
\paragraph{Models} We selected Gemini-1.5-flash from the Gemini LLM family due to its effectiveness in generating high-quality requirements user stories from concise summaries with higher emphasis on ethical aspects than O1-mini and Llama 3.1 70b~\cite{10.1145/3727582.3728689}. All parameters were set to the defaults provided by the Gemini API, except for the temperature, which was set to $0.2$ to minimize requirement creep. Thematic analysis will be conducted to map the requirements produced by the baseline and the multi-agent LLM approach.\par
\paragraph{Data} Due to the lack of publicly available datasets for benchmarking ethics requirements for AI applications, we validated our approach with two case studies from different domains: (1) \emph{Fake Review Detection System:} A web application to detect and filter fake reviews written in Arabic Language effectively, recalculating accurate product ratings for informed decision-making. (2) \emph{Saudi Sign Language App:} An app to help people with hearing impairment and communicate through Saudi Sign Language to get instant translation from short videos to textual representation form, based on work in~\cite{ALYAMI2024103774}.
\begin{figure}
    \centering
    \includegraphics[width=\linewidth]{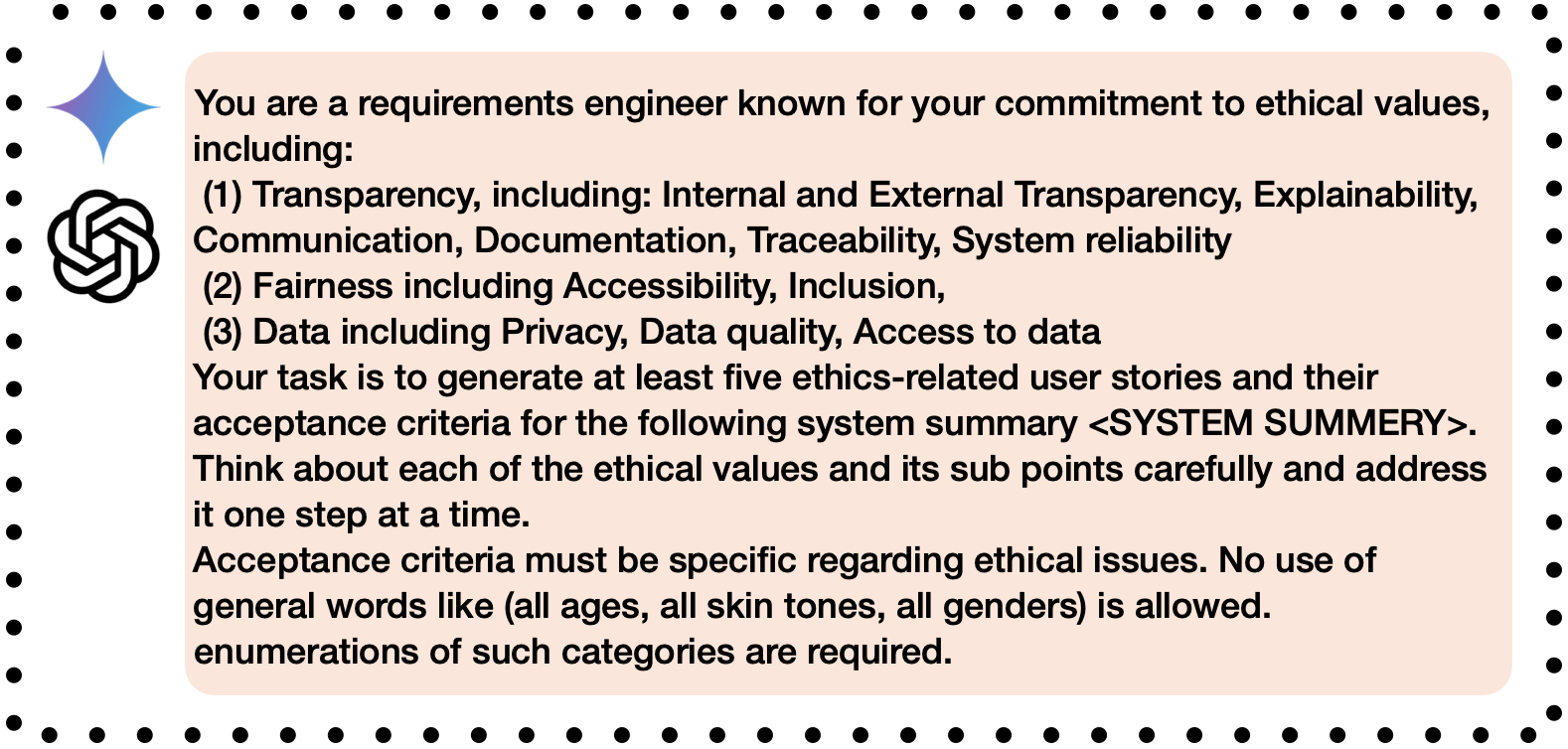}
    \caption{Single-LLM prompt.}
    \label{fig:single-LLM}
\end{figure}

\paragraph{Baselines} We compared the MALEA with human-generated and single-LLM-generated baselines:
\begin{enumerate}
    \item A domain expert or researcher with 1–3 years of expertise in the topic was interviewed for each case study. The researchers were asked to provide a summary of the system before the session in Appendix~\ref{sec:case1}. Then, using the ECCOLA framework~\cite{Vakkuri2021}, the researcher generated ethics requirements based on 12 cards under the themes of transparency, data, and fairness. Topics included internal and external transparency, explainability, communication, documentation, traceability, system reliability, accessibility, inclusion, privacy, data quality, and access to data. Each interview session lasted 30–40 minutes, and the elicited requirements were shared with the interviewee for final approval. Human-elicited requirements are in Appendix~\ref{reqs}.
    \item Requirements were generated using Google's Gemini-1.5-flash LLM~\cite{geminiteam2024geminifamilyhighlycapable}. All parameters were set to the default parameters provided by the Gemini API. Figure ~\ref{fig:single-LLM}, illustrates the prompt used for the single LLM baseline.

\end{enumerate} 

\paragraph{Mapping and Metrics} As the LLMs were instructed to generate user stories along with accompanying acceptance criteria based on the provided system description~\cite{10.1145/3727582.3728689}, for subsequent analysis, Gemini-2.5-pro was used to systematically translate these outputs into discrete requirements, treating each acceptance criteria as independent or multiple independent requirements statements when appropriate. The first author then reviewed the mapping and made any necessary corrections. A thematic analysis was conducted to map the discrete requirements to human-elicited requirements to enable their evaluation for recall compared to human-elicited requirements and the number of additional relevant requirements captured by LLMs that were missed in the human elicitation session. Precision was also reported for completeness.

\section{Results}
\label{sec:results}
In this section, we present the results of our two case studies to answer our research questions.
\begin{table*}[ht]
\centering
\caption{Evaluation results for both systems: Saudi Sign-Language (SSL) Translation = 1, Fake-Review (Fake-Rev.) Detection = 2.
\textbf{Key}: \textbf{Prod.}\,= total requirements produced; 
\textbf{TP}\,= true positives; 
\textbf{FP}\,= false positives; 
\textbf{TP\_A}\,= unique ground-truth requirements covered; 
\textbf{FN\_A}\,= ground-truth requirements missed; 
\textbf{Prec.}\,= precision; 
\textbf{Recall}\,= recall.}
\begin{tabular}{@{}llrrrrrrccc@{}}
\toprule
\textbf{System} & \textbf{Set} & \textbf{Prod.} & \textbf{TP} & \textbf{FP} & \textbf{TP\_A} & \textbf{FN\_A} & \textbf{Prec.} & \textbf{Recall} & \textbf{Unique} & \textbf{Unique and relevant} \\ \midrule
1 (SSL) & Single LLM & 24 & 12 & 12 & 8 & 5 & 50.0\,\% & 61.5\,\% & 1 & 1 \\
              & MALEA & 28 & 12 & 16 & 7 & 6 & 42.9\,\% & 53.8\,\% & 13 & 12\\ \addlinespace
2 (Fake-Rev.) & Single LLM & 21 & 9  & 12 & 6 & 2 & 42.9\,\% & 75.0\,\% & 7 & 4\\
              & MALEA & 25 & 18 & 7  & 7 & 1 & 72.0\,\% & 87.5\,\% & 4 & 4\\ \bottomrule
\end{tabular}
\label{tab:merged}
\end{table*}

\subsection{RQ1. Coverage of Human-Elicited Requirements} 
The results of the coverage of human-elicited requirements by both the single LLM and the MALEA are in Table~\ref{tab:merged}.
 \paragraph{Fake-Review Detection System} The 30-minute interview with the researcher participant, who has one year of experience developing the system, identified \emph{eight ethics requirements} in the areas of transparency, fairness, and data.  
 Twenty-one requirements were generated by the single LLM across five user stories, each including 3–4 acceptance criteria, whereas MALEA produced \emph{25 requirements} across seven user stories, each including 1–4 acceptance criteria. The baseline single LLM and MALEA achieved 75\%, and 87\% recall, respectively. \par
 Of the eight ethics requirements elicited during the interview, \emph{one} was not reproduced by either MALEA or the single-LLM baseline approach:
\begin{enumerate}
  \item \emph{``In case an expression has multiple meanings in different Arabic dialects, it shall be ignored.''}  
        Although both LLM configurations emphasized fairness across dialects, neither explicitly addressed this rule for handling polysemous expressions.
\end{enumerate}
Also, we notice that the single LLM generated a reliability requirement focused on overall accuracy, omitting precision indicated in the human-elicited requirement: \emph{``The system shall have a precision of 95\,\%.''} . Whereas, MALEA did mention precision, but only as one of several metrics: \emph{``The system achieves at least 95\,\% accuracy across various Arabic dialects and writing styles, measured using F1 score, precision, and recall.''} It also added an accountability clause \emph{``Quarterly reports track and analyze false-positive rates, identifying trends and contributing factors''}, which reinforces the precision target indirectly.

\paragraph{Saudi Sign Language Translation} The 35-minute interview with the researcher participant, who has three years of experience developing the system, identified \emph{13 ethics requirements} in the areas of transparency, fairness, and data. Twenty-four requirements were generated by the single LLM across five user stories, each including 3–6 acceptance criteria, while requirements provided by MALEA amounted to \emph{28 requirements} across 20 user stories, each including one acceptance criteria. The single LLM baseline and MALEA achieved 62\% and 54\% recall, respectively. \par
 
Of the thirteen requirements elicited from the participant, \emph{four were not reproduced} by either MALEA or the single-prompt baseline:
\begin{enumerate}
  \item \emph{``The system must respond within 5~seconds 95~\% of the time.''} under the topic of reliability.
  \item \emph{``The system developer must document any trade-off decisions made regarding resources and video length.''} under the topic of documentation.
  \item \emph{``The system shall provide a log of all mistranslated instances.''} under the topic of traceability
  \item \emph{``All participants in the development data collection must sign a consent form.''} under the topic of privacy and data
\end{enumerate}
Reporting confidence intervals were not mentioned by the single LLM baseline, whereas informing the user if it cannot process the video due to connection issues and informing the user that the video will be transferred to the cloud for processing are both not mentioned by the multi-agent LLM.
\subsection{RQ2. Added ethics requirements Beyond Human Elicited Requirements}
Large Language Models synthesize vast and diverse information, enabling them to surface potential requirements, especially nuanced ethical considerations, that might be overlooked during time-constrained human discussions like a 30-minute interview. As shown in Table~\ref{tab:merged}, reported as FP, both single LLMs and MALEA introduced ethics requirements beyond human-elicited requirements. Some of those requirements were shared by both, MALEA and the single LLM baseline; we report below on the unique results introduced by each approach.\par

 \paragraph{Fake-Review Detection System} The \emph{Single LLM Baseline} introduced 7 unique requirements. Three of which were irrelevant related to the fairness of the sentiment analysis component. While fairness in sentiment analysis is an important aspect, its relevance is limited when fairness for the overall system is already emphasized. The other four unique and relevant requirements are related to the fact that the system shall ensure that the appeal process is clearly documented in both Arabic and English and that all related content, including messages, explanations, and UI elements, is accessible, meeting WCAG AA standards, with screen reader compatibility and alternative text for visual elements. \par
\emph{MALEA} introduced four unique requirements focusing mainly on the themes of data and transparency that we can summarize with: The system shall ensure strong data protection and reliability by implementing encryption, access controls, annual penetration testing, and intrusion detection, along with defined data retention and secure deletion policies. It shall track and analyze false positives using user appeals and validated datasets and maintain 99.9\% uptime with robust error handling, logging, and regular backups. Additionally, for the six requirements, more specific details were provided to make them testable or less ambiguous. These included listing metrics for bias assessment, such as statistical parity and equalized odds with $p < 0.05$, and specifying a 72-hour response timeline for appeals of incorrectly flagged fake reviews.\par

 \paragraph{Saudi Sign Language Translation} The \emph{Single LLM Baseline} introduced only one unique requirement related to reporting a real-time reliability metric (e.g., translation accuracy rate over the past hour) for monitoring the system. \par
\emph{MALEA} introduces 12 unique and relevant requirements across the different themes, which can be summarized as follows: The system shall ensure timely feedback responses, rigorous bias detection and mitigation evaluation, and consistent accuracy under varied conditions. It shall implement strong privacy measures, including audited anonymization, opt-out options, and limited data use for model improvement. Additionally, features for diverse cognitive/sensory abilities and establishing an incident response plan. Furthermore, some requirements were enhanced with more specific or testable details. These enhancements included specifying feedback mechanisms with response time thresholds, subgroup disparity thresholds, and encryption protocol names.
The less relevant requirement is related to 'The system must implement ISO 27001 controls and verify through regular security audits,' which, although a valuable step towards better information security, may not be necessary depending on the contract or law under which the system operates. 

\section{Discussion}
\label{sec:discussion}
In this section, we discuss some key takeaways from the empirical evaluation for future consideration.

\subsection{Reliability of the Generation}
The use of LLMs for generating ethics requirements shows promising results for creating initial ethics requirements drafts. They can even capture, sometimes, cultural and context-specific ethical nuances, as observed in our second case study, where a single LLM approach response recognized the different garments worn by women in Saudi Arabia: \emph{``The system's training data includes diverse representations of Saudi Sign Language users, including women wearing hijabs, niqabs, and other culturally appropriate attire. Specific numbers of training samples for each category should be specified, if possible, e.g., at least 100 videos of women wearing hijabs, at least 50 videos of women wearing niqabs, etc."} However, this cultural dimension was not captured consistently when the prompt later was repeated with a different seed, nor by the multi-agent approach in the analyzed run (the first run) indicating reliability concerns. 

\subsection{Ambiguity Reduction} 
Using the multi-agent collaboration to refine the requirements, MALEA consistently supplements qualitative statements with quantitative thresholds that testers can verify. For example, while the single LLM does state that \emph{``All video data shall be encrypted both in transit and at rest"}, MALEA specifies TLS 1.3 or a higher standard at transmission and AES-256 encryption at rest. Another example is the appeal process in which MALEA specified a response window of 72 hours for any appeal. MALEA leaves a [PLACEHOLDER] tag that signals to reviewers exactly where clarification is required (e.g., “statistical parity gap $\leq$ Z \%”) when it lacks the details. In addition, MALEA agents looked more closely at the entire software cycle, including maintenance such as incident response plans, transparency reports, quarterly public KPIs, and rollback documentation of bias mitigation, none of which surfaced in the single-LLM output.

\subsection{Human-in-the-loop Integration}
\emph{MALEA} enriches initial LLM drafts by inserting relevant guidelines, policies, metrics, and thresholds, but does not replace stakeholder collaboration. By embedding explicit placeholders or a variable for human review, the system supports an iterative elicitation process. Requirements engineers and domain experts should replace each placeholder with actual values to ensure completeness and contextual accuracy. This human oversight is essential for capturing cultural nuances and filling context-specific gaps that automated generation alone cannot address. These placeholders compel requirements engineers to engage with domain experts rather than rely solely on LLM outputs, thereby preserving human autonomy and ensuring responsible integration of ethical considerations.

\subsection{Practical Implications for RE Teams}
\emph{MALEA} enables faster first drafts of ethics requirements by allowing RE teams to invoke the system immediately after scoping a new feature or module using the description only as input. Instead of conducting a large, multi-stakeholder workshop solely to generate initial ethics-related user-story candidates, engineers receive a draft set of requirements in minutes with placeholders to further expand the discussion. As this proposed approach is designed to \emph{complement, not replace, existing frameworks} such as ECCOLA or RE4HCAI, it enables RE teams and stakeholders to focus on prioritization, tradeoff analysis, validating LLM-generated drafts and requirements with greater ambiguities.

\subsection{Limitations and Threats to Validity}\label{sec:threats}

This study’s findings should be interpreted in light of the following limitations and validity threats.

\paragraph{Construct Validity}  
The mapping of LLM-generated user stories to discrete requirements was performed by LLMs and then reviewed by the first author. While this process followed a consistent procedure, it still involved subjective judgment, particularly in interpreting some acceptance criteria as standalone requirements, and others were split into multiple standalone requirements. 

\paragraph{Internal Validity}  
Large language models (LLMs) can exhibit nondeterministic behavior: different random seeds or prompt variations may yield divergent outputs. In this evaluation, we evaluated only a single seed per case study, so the reported recall and added‐value metrics may not generalize across runs. Moreover, in the current prompt, we only state the ECCOLA's card topics without including the motivation, questions, and examples in a zero-shot setting. Hence, the depth and interpretation of such a topic could vary by run. This should be investigated in further iterations of this approach. Additionally, inherent biases present in LLM training data may propagate into the proposed ethics requirements, leading to incomplete or skewed recommendations. We address this by emphasizing the importance of human‐in‐the‐loop validation.

\paragraph{External Validity}  
Our evaluation is limited to two AI applications (fake‐review detection and sign‐language interpretation) and three ethics‐theme categories (transparency, fairness, and data). As a result, generalizability to other domains (e.g., medical AI, autonomous vehicles) and to the full spectrum of AI ethics guidelines (e.g., sustainability, human autonomy, security) is untested. Broader validation with diverse stakeholders and additional case studies or on a benchmarking dataset is recommended to establish the approach’s robustness across contexts.

\paragraph{Conclusion Threat}  
Because the human gold sets were produced by a small number of domain experts during short ECCOLA sessions, the reference requirements may themselves be incomplete. This “ground‐truth” could omit relevant ethics requirements affecting precision measurements. Future studies should involve larger, more diverse expert panels to strengthen the validity of baseline comparisons.

A general limitation of this approach is the potential risk that organizations may adopt automated ethics‐elicitation tools purely to satisfy regulatory or marketing objectives, \emph{ethics‐washing}, without genuine stakeholder engagement. Embedding explicit placeholders and mandatory human review partially mitigates this risk, but institutional incentives and governance practices are also required to ensure meaningful ethical deliberation.

\subsection{Future directions}
There are many open problems and directions that are open for research and improvement.
\paragraph{Parallelization} While the current proposed approach provides depth coverage through the multi-agent conversation, breadth coverage through parallelization would be an interesting approach to investigate, with each ethics advocate advocating for a different persona that covers appropriate stakeholders according to~\cite{10.1145/3514094.3534187} along with its ethical preferences may help mitigate the issue of stochasticity in LLMs, which have a higher chance of covering all perspectives.
\paragraph{Comprehensive evaluation} As this task lacks a development dataset necessary to prompt tune LLMs or benchmarking datasets to evaluate such an approach, a benchmarking dataset is needed that includes diverse stakeholders for diverse systems to measure the alignment of LLMs in requirements engineering with human values.
\paragraph{Human-in-the-loop support} Currently, placeholders are put in place to request stakeholder's engagement. However, LLMs were not instructed on when to put them and when to put actual values, which is an important point to investigate. Moreover, with the advancement of tooling in multi-agent LLMs, conversation between LLM agents and non-technical stakeholders could be an interesting direction to investigate to resolve ambiguities and provide more domain knowledge when needed.
\paragraph{Integration with existing frameworks and AI ethics guidelines} When we crafted our ethics advocate system prompt, we only considered three themes of the ECCOLA framework~\cite{Halme2021} along their subtopics without including their definitions or related questions. Incorporating RAG techniques could enable customizing the MALEA to attain different guidelines and legal documents without issues related to prompt lengths. Moreover, faithfulness analysis should be conducted to measure how faithful the MALEA is to the specified guidelines. 
\paragraph{Prioritization} While a large number of ethics-related requirements could be produced by our proposed approach, prioritization techniques should be implemented to avoid requirements creep. This should take into account stakeholders' value systems and context.
\paragraph{Ablation studies regarding LLM choice} While the selection of the LLM in this study was derived from the literature, more recent models from the same LLM family have emerged, and other powerful LLMs across multiple LLM families. An ablation study for different LLMs would be interesting to investigate whether more advanced (or simpler) LLMs would affect the recall and additional relevant ethics requirements. 

\section{Conclusion}
\label{sec:conclusion}
This research explores the potential of multi-agent large language model systems to automate the elicitation of ethics requirements early in the software development lifecycle. Through a preliminary evaluation involving two distinct AI applications, our proposed multi-agent approach demonstrated significant effectiveness. Across both systems, MALEA demonstrated higher performance, achieving a recall of 81.08\%. This surpassed the single LLM approach, which yielded a recall of 75.0\%. In addition, it introduced additional relevant ethics requirements for consideration and systematically addressed quality concerns by adding placeholders and relevant metrics and standards. However, inherent reliability limitations necessitate continued human involvement and oversight when using MALEA. This work makes a substantial contribution to integrating ethical considerations systematically into requirements engineering, thereby promoting the development of ethically robust AI systems.

\section*{Acknowledgment}
The authors would like to acknowledge the support received from Saudi Data and AI Authority (SDAIA) and King Fahd University of Petroleum and Minerals (KFUPM) under SDAIA-KFUPM Joint Research Center for Artificial Intelligence Grant no. JRC-AI-RG-12. The authors would also like to thank Sara Alyami and Reem Aljunaid for their involvement in the case studies.

\bibliography{custom}
\bibliographystyle{ieeetr}

\appendix

\subsection{System Summaries}
\label{sec:case1}
\textbf{Case study 1: Arabic Fake Review Detection System}
Fake reviews can distort online product ratings, mislead consumers, and harm platform integrity. To address this, our system employs advanced AI and NLP methodologies to detect and filter fake reviews effectively, recalculating accurate product ratings for informed decision-making.
The detection mechanism uses a comprehensive feature set, including:
1- Sentiment-Based Features: Sentiment polarity (positive, negative, or neutral). Counts of positive and negative words. Sentiment inconsistencies, such as mismatches between review ratings and textual sentiment or deviations from average product/user sentiment.
2- Text-Based Features: Sentence, word, and character counts. Tip words, emails, and link frequencies. Categorization of text length (e.g., very long, medium, short). Cosine similarity for textual content across reviews, highlighting potential duplication or unnatural patterns.
3- Behavioral Features: User-specific anomalies, such as maximum reviews per day or deviation from average ratings. Review count distributions, including those with specific ratings or inconsistencies.\par

\textbf{Case study 2: Saudi Sign Language (SSL) System} Saudi Sign Language is a set of hand gestures and symbols used by the Deaf in Saudi Arabia. It is also used by people of normal hearing to communicate with the deaf. This system is for a mobile app that provide real-time translation for people using Saudi Sign Language. It leverages artificial intelligence to translate sign language sentences from video into corresponding text representations. Unlike isolated sign recognition, the system processes continuous sign sequences without explicit boundaries between signs, making it more complex. The app is used for capturing the video and displaying the translation, whereas all the processing is done in the cloud. The system can also be used as a web application to be used in help desks across different institutions.
This system plays a vital role in facilitating the integration of Deaf individuals into hearing communities by bridging the communication gap between the two. By removing language barriers, CSLR systems empower Deaf individuals to participate more actively in education, workplaces, and social environments, while also raising awareness and promoting greater acceptance of sign languages among hearing populations.
\subsection{Human elicited ethics requirements}
\label{reqs}
\textbf{Arabic Fake review detection}
1- The system shall be transparent to product owners and website administrators on how the decisions are made.
2- The system shall indicate that the rating of product is adjusted by AI to remove fake reviews
3- The system should implement a mechanism to report the incorrect labeling as fake review.
4- The system shall have a precision of 95\%
5- The development of the system must be documented to increase trust in the system
6- The system shall support both dialectal Arabic and MSA Arabic
7- The training and testing data shall be representative of both dialectal Arabic and MSA Arabic.
8- In case of an expression having multiple meanings in different Arabic dialects it shall be ignored.

\textbf{SSL System}
1- Detail of the system shall be transparent to future maintainers.
2- The system shall inform the user that the video will be transferred to the cloud for processing.
3- The system shall inform the user that the translation is provided via artificial intelligence.
4- The system developer must document any trade-off decision made regarding the resources and the length of the video
5- The system shall enable the users to receive feedback on the translation
6- The system shall provide a log of all mistranslated instances
7- The accuracy of the system must exceed 95\%
8- The system must respond within 5 seconds 95\% of the time.
9- The system must inform the user if it cannot process the video due to connection issues.
10- All participants in the development data collection must sign a consent form.
11- The system shall inform the user if the confidence of the generated translation is below 90\%
12- All data must be sent and received in an encrypted manner
13- The system should maintain high accuracy for all users including various age groups, body built, skin color, genders, people with eyeglasses, people with gloves, people with face masks, and people with niqab

\end{document}